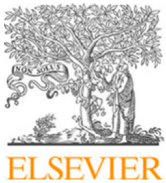
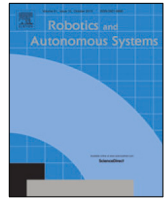

# Improving robot dual-system motor learning with intrinsically motivated meta-control and latent-space experience imagination

Muhammad Burhan Hafez [*], Cornelius Weber, Matthias Kerzel, Stefan Wermter

*Knowledge Technology, Department of Informatics, University of Hamburg, Germany*

**ABSTRACT**

Combining model-based and model-free learning systems has been shown to improve the sample efficiency of learning to perform complex robotic tasks. However, dual-system approaches fail to consider the reliability of the learned model when it is applied to make multiple-step predictions, resulting in a compounding of prediction errors and performance degradation. In this paper, we present a novel dual-system motor learning approach where a meta-controller arbitrates online between model-based and model-free decisions based on an estimate of the local reliability of the learned model. The reliability estimate is used in computing an intrinsic feedback signal, encouraging actions that lead to data that improves the model. Our approach also integrates arbitration with imagination where a learned latent-space model generates imagined experiences, based on its local reliability, to be used as additional training data. We evaluate our approach against baseline and state-of-the-art methods on learning vision-based robotic grasping in simulation and real world. The results show that our approach outperforms the compared methods and learns near-optimal grasping policies in dense- and sparse-reward environments.

## 1. Introduction

Reinforcement learning (RL) enables artificial agents to learn how to perform sequential decision-making tasks from experience, without manually programming the desired behavior. This involves online learning of a control policy – a mapping from a raw and often high-dimensional sensory input to a raw motor output that optimizes the task performance. In recent years, deep RL has been used to learn this mapping from self-collected experience data by utilizing deep neural networks as function approximators, achieving superhuman performance in a variety of game domains [1,2] and facilitating the acquisition of complex robotic manipulation skills [3,4].

The current success of deep RL, particularly in robotic control, has come at the expense of notoriously high sample complexity that fundamentally limits how quickly a robot can learn successful control policies. A wide range of approaches have been proposed to alleviate this problem. Some focused on improving experience replay in which data are drawn from a memory of recent experiences and used for training the controller. Sampling experiences with a probability proportional to reward prediction error instead of random sampling [5] and counting unsuccessful policy rollouts as successful ones by replaying experiences with a different goal than the one the agent was trying to achieve [6] are two prominent examples. Others proposed extending count-based exploration methods, previously only applicable to tabular representations, to problems with continuous high-dimensional state spaces [7,8]. Learning a separate exploration policy was also found to increase the sample efficiency of learning the target policy by updating the exploration policy based on the amount of improvement in the target policy as a result of the experience data collected with the exploration policy [9]. Similarly, the learning speed on novel tasks was found to be improved by using a task-independent exploration policy updated between learning trials of different tasks [10]. In a different class of approaches, intrinsic rewards were used to efficiently guide the exploration in environments where extrinsic rewards are sparse or absent. Proposed intrinsic reward functions include novelty estimation of perceived states [11,12], learning progress in predicting future states [13], competence progress in achieving self-generated goals [14,15], and information-theoretic measures of uncertainty [16,17].

Despite the efficiency gained by these approaches, learning the desired control behavior was predominantly model-free. While a few approaches (e.g., [11,13]) learn transition models, the models were only used for computing auxiliary feedback signals to improve exploration. The policy function itself was learned with model-free RL. This places a limit on the achievable degree of sample efficiency and is inconsistent with a large body of behavioral and neural evidence showing that model-free and model-based learning systems both have an active role in human motor learning [18,19].

* Corresponding author.
 *E-mail addresses:* hafez@informatik.uni-hamburg.de (M.B. Hafez),
weber@informatik.uni-hamburg.de (C. Weber),
kerzel@informatik.uni-hamburg.de (M. Kerzel),
wermter@informatik.uni-hamburg.de (S. Wermter).

In the following, we will discuss in more detail dual-system motor learning and how it is applied in robotics. We will then introduce a new approach that integrates the two learning systems in an adaptive, reliable and sample-efficient manner and perform grasp-learning experiments where we implement and evaluate our approach on a humanoid robot.

*1.1. Dual-system motor learning*

Motor behavior can be divided into habitual behavior obtained by model-free learning and goal-directed behavior obtained by model-based learning [20]. Several hypotheses were proposed to explain how the human brain arbitrates between model-based and model-free learning systems. For instance, Cushman and Morris [21] argue that when performing a sequential decision-making task, humans use the model-free system to habitually select goals and then the model-based system to generate a plan to achieve a selected goal. Another study proposes a contrasting hypothesis called "plan-until-habit", in which planning is first performed by simulating the world up to a certain depth that decreases with increased time pressure and then model-free action values are exploited [22]. While this study attributes the change in behavior between model-based and model-free control to the availability of cognitive resources, particularly time, other studies have found that the behavior instead changes according to the expected reward regardless of resource availability [23,24]. Kool et al. [23] support the latter by providing behavioral evidence that people with a perfect transition model of the task and an extended response deadline exerted less model-based control when its expected reward advantage was lower than the cognitive cost involved. This finding was interpreted by suggesting that the brain estimates the value of using each control system but reduces that of the model-based system in proportion to its increased cognitive cost. Similarly, Boureau et al. [24] state that meta-decisions including arbitration between model-based and model-free control are governed by a cost-benefit trade-off in which the brain constantly generates rough estimates of the costs and benefits of allocating cognitive resources for model-based control. The average reward rate, which is the reward expected for temporally allocating a particular resource, and the controllability, which measures how advantageous a carefully considered decision is over a fast habitual one in terms of rewards collected, are proposed as estimates for the opportunity cost and benefit respectively. The willingness to exert model-based control thus increases proportionally to how much larger the reward obtained by controllability is compared to the average reward rate. The authors note, however, that modifying these estimates by including other decision variables like the uncertainty about action outcomes might account for meta-decisions in specific behavioral contexts like exploration.

Haith and Krakauer [18] review the behavioral evidence for the existence of each of the model-based and model-free mechanisms of motor learning in humans and argue that both are employed in parallel by the motor system for movement control. They point out that while the two learning systems generate their own estimate of the value of a given action at a given state, the decision which action to take is made primarily based on the reliability of each of these two estimates. Imperfect predictions of an internal forward model limit the reliability of model-based learning and, hence, in the later stages of learning, after extensive experience, model-free learning becomes more reliable, as the authors indicate. Another study provides neural evidence that the human brain encodes the reliability of model-based and model-free learning systems based on their prediction errors in the lateral prefrontal and frontopolar cortex and uses the reliability signals to dynamically arbitrate behavioral control between the two systems [19]. The arbitration model the study proposes combines model-based and model-free value signals, weighted by the degree of reliability of each system, and uses this integrated value signal for guiding behavior. To account for the cognitive complexity involved, the arbitrator incorporates a bias toward the less cognitively demanding model-free control. The arbitration between different learning systems was also found to drive human strategy selection where the goal is to learn when to use which strategy [25]. The proposed context-sensitive strategy selection approach, which assumes a mental model predicting each strategy's accuracy and execution time from features of the current situation, was found to better explain how people adaptively choose strategies than previous accounts. However, it is based on choosing the strategy with the best predicted speed-accuracy trade-off rather than choosing the most reliable strategy.

In contrast to previous works suggesting strict neural and behavioral division between model-based and model-free learning systems, Russek et al. [26] propose a computational framework where the two systems are tightly coupled, motivated by evidence supporting a role for dopamine in model-based learning besides its well-established role in model-free learning. In the framework, action values are estimated by applying model-free temporal difference (TD) learning to successor representations (SR), which are the expected future state occupancies. This was found to explain the involvement of dopamine in model-based learning, since the TD error is a reward prediction error thought to be mediated by phasic dopamine and SR is a predictive representation capturing knowledge of the transition model. Although the presented framework gives a neurally plausible computational account of the interaction between the two learning systems, it does not answer the question of how the brain prioritizes computations and arbitrates control between learning systems, as concluded by the authors.

*Dual-system deep RL for robot control.* Deep RL approaches are broadly classified into model-based and model-free ones. Model-based approaches facilitate transfer of learning across tasks, since a model learned in the context of a task can be directly used to compute an appropriate control policy for a new task. They are also typically sample-efficient in that they allow for generating synthetic experiences by making predictions about the future. On the other hand, model-free approaches do not have representational limitations that would prevent the convergence to a desired behavior if the model representation is insufficient to perfectly capture the environment dynamics. However, they require a lot of experience and hence have high sample complexity. This has motivated several works to address the problem of how to combine the benefits of model-based and model-free methods.

Initializing the neural network policy of a model-free learner using rollouts of a model-based controller followed by model-free fine-tuning of the policy was found to lead to a higher sample efficiency compared to pure model-free learning with random policy initialization [27]. The model-based controller used is based on random-sampling where several randomly generated action sequences are fed to the model and then the sequence with the highest expected reward is chosen. This limits the effectivity of the approach to low-dimensional action spaces and short horizons. In a different work, Feinberg et al. [28] decompose action-value estimation into two parts: one contains the sum of future rewards predicted by a learned model over a limited horizon and one contains the cached model-free estimate of the long-term reward computed at the end of the horizon. While the method is shown to boost the sample efficiency, it assumes perfect model predictions for a fixed horizon, which is a strong assumption, because, in practice, the model generates noisy data early in learning, and a measure of model reliability is therefore needed.

Other works used information about the future provided by a trained world model as input to the model-free learner to improve its decisions [29,30]. In [29], imagined trajectories generated by a model are processed by a recurrent neural network that outputs a rollout encoding for each trajectory. The encodings are concatenated and used as additional context for the model-free learner's value and policy networks. Rather than training a feedforward model, [30] train a recurrent world model on random environment rollouts and use the hidden state of the trained model along with a learned abstract state representation as input to a model-free controller. The proposed approach achieves state-of-the-art performance on an image-based car racing task. However, these works employ pretrained world models and abstract representations with the risk of encoding task-irrelevant features. Instead, Francois-Lavet et al. [31] propose training the world model and an abstract state representation that minimizes both model-free and model-based losses during task learning. The abstract state is the input to both the model-free Q-network predicting action values and the world model predicting next states and rewards. Planning is done by performing one fixed-depth rollout of the model for each possible action at the current state and then taking the first action of the rollout with the highest overall estimated value. The approach has two major drawbacks: first, the complexity of planning, which is performed at each time step, grows exponentially with the number of possible actions; second, if the model is inaccurate, as is the case in complex domains, a large fixed planning depth leads to a compounding of prediction errors that eventually impair task performance.

Another line of research focused on modeling an arbitrator that explicitly switches control between model-based and model-free systems at decision time [32,33]. One study proposes a control architecture where the arbitrator chooses between an action suggested by an inverse dynamics model and an action suggested by an actor network of a model-free actor–critic system [32]. The arbitration is guided by reward prediction error. If the error at the previous time step is below a predefined threshold, the actor network's prediction is performed. Otherwise, the inverse model's predicted action is performed. The approach does not address model imperfection and is evaluated on a robotic reaching task with a significantly low-dimensional state space. In contrast, an estimate of reliability of a learned model's predictions was recently used to guide the arbitrator's decision which of the two systems to query for an action during robot grasp learning [33]. The reliability is measured by the model learning progress, which is the time derivative of the average prediction error of the model. If the reliability is positive, the arbitrator queries the model-based system for an action, which performs gradient-based model predictive control, and if not, the arbitrator queries the model-free system instead. The approach, however, relies only on the temporal information when computing the learning progress and ignores the spatial context. It also uses a fixed planning horizon.

*1.2. Experience imagination in dual-system motor learning*

One complex cognitive process the dual-system motor learning implements is experience imagination, using the capacity of the world model to make predictions about future states and rewards. Imagination typically refers to the mental simulation of motor behavior. It requires a combination of different cognitive functions, including episodic memory, abstract sensory and motor representations, and manipulation of representations [34]. This cognitive synergy the imagination induces was distinguished neurally by identifying the different brain regions activated during imagination, including both cognitive and motor areas [35,36], and is strongly associated with the cognitive development in children where complex behaviors develop from the recombination of simple ones. Experience imagination is also essential to mental practice and cognitive rehearsal of physical skills, facilitating skill acquisition [37]. Besides, it is estimated that automating imagination has the potential of advancing deep learning beyond finding correlations in data as well as providing a means to broaden the focus of research from problem solving to problem creation through the imagination-supported ability to self-generate goals and pursue them with intrinsic motivation [38,39].

Based on how experience imagination is implemented, dual-system deep RL approaches can be divided into two groups: (i) online imagination [27–33] and (ii) offline imagination [40–42]. In online imagination approaches, generating imagined trajectories for planning with the world model is done at decision time, as discussed in Section 1.1. Offline imagination approaches, on the other hand, augment the memory of real experiences with model-generated imagined experiences, increasing the amount of training data available to the RL agent when updating the control policy offline with experience replay. Gu et al. [40] apply offline imagination by generating on-policy imagined rollouts under the model, starting at states sampled from transitions the model has recently been trained on, and adding them to the replay memory. While this results in a fast convergence to an optimal policy with model-free RL and shows some robustness to imperfectly learned models, the used linear model is insufficient to perfectly capture complex environment dynamics and generate correct imagined rollouts in tasks involving learning from high-dimensional observations, as indicated by the authors. Instead of always using imaginary data in training, a different work suggests to update the policy and value networks from imagined rollouts only when there is a high uncertainty in the estimated action values [41]. The approach is shown to achieve high sample efficiency on continuous control tasks but does not consider model prediction errors. In visuomotor control tasks, generating imagined data requires learning perfect world models at the pixel level, which is impossible in practice. This has recently been addressed by learning the model in latent space and dividing the experience replay buffer into pixel-space and latent-space buffers for storing real and imagined experiences respectively [42]. In the approach, the learned latent space is self-organized into local regions with local world models, and a running average of the model prediction error is independently computed for each region. Unlike previous works, the imagined rollout is reliably generated with a probability inversely proportional to the average error of the current region, and the imagination depth is adaptively determined by the average error of the traversed regions.

*1.3. Issues with model-based planning and proposed changes*

As discussed in Section 1.1, recent works on dual-system deep RL have demonstrated the potential of using model predictive control (MPC) for planning [27,31,33]. MPC is an iterative optimization-based control method that collects a multi-step rollout from an initial state given a dynamics model, infers an optimal action plan, performs the first action of the plan, and then repeats the process in a receding horizon. The time and space complexity of planning with MPC is very high when performing backpropagation through time to optimize the MPC planning objective, particularly over long planning horizons. Amos et al. [43] attempt to address this issue by implicit differentiation of the Karush–Kuhn–Tucker (KKT) optimality conditions at a fixed point of the employed convex optimization solver. In the approach we present here, a different solution is proposed by arguing that the length of the planning horizon should always be adaptively determined according to the current reliability of the learned model of environment dynamics. Another issue is the inevitable model errors that quickly compound during multi-step planning. Forcing latent variables to predict the long-term future using an auxiliary cost during model training was found to make planning in the latent space involve less prediction errors [44]. In our approach described in Section 2, we instead focus on developing a directed exploration strategy that gradually improves the model accuracy.

In light of the previous research, we make the following contributions:

- First, we present a novel dual-system motor learning approach where an intrinsically motivated meta-controller arbitrates online between model-based and model-free decisions based on the local reliability of a learned world model.
- Second, we describe a new learning framework that integrates online meta-control with offline learning-adaptive experience imagination.
- Finally, we show that our proposed framework improves the sample efficiency of learning vision-based motor skills on a developmental humanoid robot, compared to baseline and recent dual-system methods.

## 2. Intrinsically motivated meta-control

We describe here our approach to dual-system motor learning. The approach consists of model-free and model-based control systems and a meta-controller deciding which of the two systems to query for an action at each time step. We first present the two systems and then discuss how the local reliability in model predictions is used to adaptively guide meta-decisions and provide intrinsic feedback to improve the learned model. Our objective is to train a policy neural network representing the desired control behavior more efficiently than when following a pure model-based or model-free approach.

### 2.1. Model-free control system

To train a model-free controller from experience, we consider a standard model-free reinforcement learning (RL) problem where the goal is to learn a policy, $\pi : S \to P(A)$, mapping from states $S$ to probability distributions over actions $A$ that maximizes the expected return $R_t = \sum_{i=t}^{T-1} \gamma^{i-t} r(s_i, a_i)$ under $\pi$, where $r : S \times A \to \mathbb{R}$ is the reward function and $\gamma \in [0, 1]$ is the discount factor. In RL, actor–critic methods are well suited for continuous control by learning simultaneously a value function and a policy function. We are particularly interested in off-policy actor–critic methods, since they allow for learning from actions coming from different systems, such as a model-based controller. Deep Deterministic Policy Gradient (DDPG) [45] is a state-of-the-art off-policy actor–critic method that we use in our approach along with an off-policy variant of Continuous Actor–Critic Learning Automaton (CACLA) [46].

The action-value ($Q$-)function is defined as the expected return of taking a particular action at a particular state and following a policy $\pi$ thereafter: $Q^\pi(s_t, a_t) = \mathbb{E}[R_t|s_t, a_t]$. Accordingly, the optimal policy $\pi^*$ satisfies $Q^{\pi^*}(s, a) \geq Q^\pi(s, a), \forall (s, a) \in S \times A$. In infinite or large state–action spaces and when the transition model is unknown, actor–critic methods approximate the $Q$-function and the policy function using critic $Q(\cdot, \cdot|\theta^Q)$ and actor $\mu(\cdot|\theta^\mu)$ neural networks parameterized by $\theta^Q$ and $\theta^\mu$ respectively. The critic is trained to minimize the loss between the current value estimate $Q(s_t, a_t|\theta^Q)$ and the target value $y_t = r_t + \gamma Q'(s_{t+1}, \mu'(s_{t+1}|\theta^{\mu'})|\theta^{Q'})$, where $Q'$ and $\mu'$ are the target critic and actor networks parameterized by $\theta^{Q'}$ and $\theta^{\mu'}$ respectively and updated slowly towards their corresponding $Q$ and $\mu$ networks:

$$\mathcal{L}_Q(\theta^Q) = \left(y_t - Q(s_t, a_t|\theta^Q)\right)^2. \tag{1}$$

The actor, however, is trained differently by each method. DDPG updates the actor's parameters by minibatch gradient ascent on the $Q$-function or, equivalently, minibatch gradient descent on the loss:

$$\mathcal{L}_{DDPG}(\theta^\mu) = -\frac{1}{n}\sum_i Q\left(s_i, \mu(s_i|\theta^\mu)|\theta^Q\right), \tag{2}$$

where $n$ is the minibatch size. The gradient of $\mathcal{L}_{DDPG}$ with respect to the actor parameters is called the policy gradient: $\nabla_{\theta^\mu}\mathcal{L}_{DDPG}(\theta^\mu) = \frac{1}{n}\sum_i \nabla_a Q(s, a|\theta^Q)|_{s=s_i, a=\mu(s_i)} \nabla_{\theta^\mu}\mu(s|\theta^\mu)|_{s=s_i}$. Off-policy CACLA, on the other hand, updates the actor only when the advantage $\delta_t$ of taking the current action $a_t$ is positive by minimizing the loss:

$$\mathcal{L}_{CACLA}(\theta^\mu) = (a_t - \mu(s_t|\theta^\mu))^2|_{\delta_t > 0}, \tag{3}$$

where $\delta_t = r_t + \gamma Q'(s_{t+1}, \mu'(s_{t+1}|\theta^{\mu'})|\theta^{Q'}) - Q(s_t, \mu(s_t|\theta^\mu)|\theta^Q)$ is the action advantage at time step $t$ representing how better the observed value of taking $a_t$ is than the expected on-policy value $Q(s_t, \mu(s_t|\theta^\mu)|\theta^Q)$. This moves the actor's output towards $a_t$ that has a positive advantage.

Our actor–critic architecture is shown in Fig. 1. In the architecture, a latent state representation $\phi_{(s_t)}$, the output of the encoder $f(s_t|\omega)$, is learned to be state discriminator and value predictor by jointly optimizing the combined reconstruction and value prediction loss:

$$\mathcal{L}_{combined}(\omega, \theta^Q, \tilde{\omega}) = \lambda_{rec}\, \mathcal{L}_{rec}(\tilde{\omega}, \omega) + \lambda_Q\, \mathcal{L}_Q(\theta^Q, \omega), \tag{4}$$

where $\mathcal{L}_{rec}(\tilde{\omega}, \omega) = \|g(\phi_{s_t}|\tilde{\omega}) - s_t\|_2^2$ is the reconstruction loss between the decoder's output $g(\phi_{s_t}|\tilde{\omega})$ and the original input $s_t$, $\mathcal{L}_Q(\theta^Q, \omega)$ is the value prediction loss Eq. (1) and $\lambda_{rec}$ and $\lambda_Q$ are weighting coefficients of individual loss components. Since it captures task-relevant information sufficient to reconstruct the original input and identify rewarding states, this jointly learned latent representation can then be used as a direct input to an actor network, as shown in Fig. 1(b). Any off-policy actor–critic method can be used together with our architecture, such as DDPG and off-policy CACLA.

### 2.2. Model-based control system

In our proposed approach, a predictive model of the world dynamics is learned simultaneously with the task. Instead of learning the model at pixel-level, which is noise sensitive and infeasible in practice, the model is learned in jointly trained latent space (Fig. 1(a)). This also ensures that the model is learned on task-relevant latent representations, as opposed to representations learned only to minimize the pixel-level reconstruction error of an autoencoder, which includes no information on what features are useful for the task. The latent-space world model predicts the next latent state representation and environment reward given the current representation and control action and is trained to minimize the loss:

$$\mathcal{L}_{model}(\theta^\mathcal{M}, \theta^\mathcal{R}) = \|\mathcal{M}(\phi_{s_t}, a_t|\theta^\mathcal{M}) - \phi_{s_{t+1}}\|_2^2 + \|\mathcal{R}(\phi_{s_t}, a_t|\theta^\mathcal{R}) - r_t^{ext}\|_2^2, \tag{5}$$

where $r_t^{ext}$ is the extrinsic reward from the environment, $\mathcal{M}(\cdot, \cdot|\theta^\mathcal{M})$ and $\mathcal{R}(\cdot, \cdot|\theta^\mathcal{R})$ are two feedforward neural networks for predicting the next latent state representation and environment reward respectively.

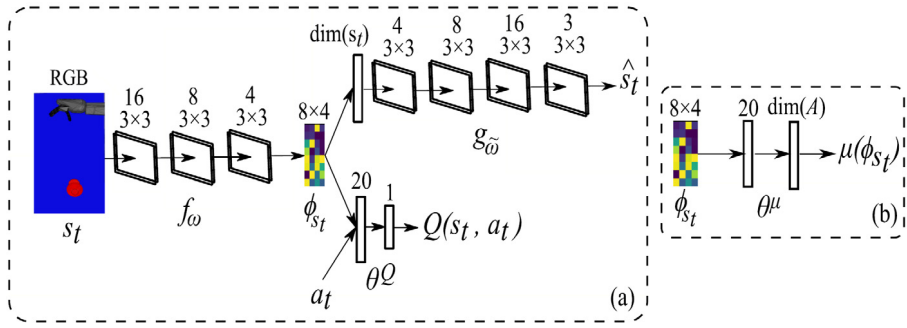

**Fig. 1.** Model-free control system: (a) Critic-autoencoder network consisting of a fully convolutional encoder $f_\omega$ that takes in a raw image $s_t$, a fully convolutional decoder $g_{\tilde{\omega}}$ that computes a reconstruction $\hat{s}_t$, and a critic $Q$ that estimates the $Q$-value given $s_t$ and $a_t$; (b) Actor network taking in the latent state representation $\phi_{(s_t)}$, which is jointly trained to minimize the reconstruction and value prediction losses, and generating a control action $\mu(\phi_{s_t})$ with a dimensionality of dim($A$), where $A$ is the action space.

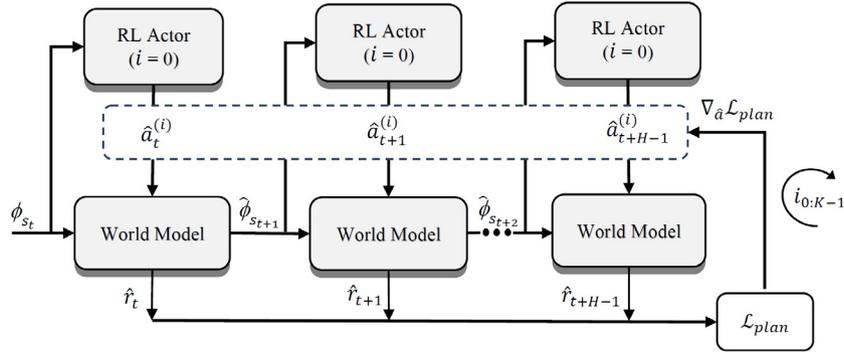

**Fig. 2.** Model-based control system: After observing the latent state $\phi_{s_t}$, the world is simulated $H$ time steps into the future using the learned world model and an action sequence $\hat{a}^{(0)}_{t:t+H-1}$ proposed by the RL actor, resulting in a sequence of model-generated latent states $\hat{\phi}_{s_{t+1:t+H}}$ and rewards $\hat{r}_{t:t+H-1}$. The objective $\mathcal{L}_{plan}(\hat{a})$ is then measured and optimized by performing backpropagation and $K$ steps of gradient descent. The first action of the optimal plan $\hat{a}^{(K-1)}_t$ is applied in the environment and the optimization process is repeated at the next time step.

To perform motor control with the latent-space world model, we use model predictive control (MPC). In MPC, the world model is rolled out multiple time steps into the future starting from an initial world state and action plan. An objective function is measured at each time step, and then by backpropagation through time and gradient descent, an action plan that optimizes the objective is computed. Only the first action of the optimal plan is taken before repeating the process again at the next time step with the updated state information in closed loop. In our approach, the initial sequence of actions is provided by the model-free RL actor (Section 2.1) at the initial and subsequent model-generated latent states and is optimized with MPC over a time horizon $H$ by minimizing the loss:

$$\mathcal{L}_{plan}(\hat{a}) = \left(R^* - \sum_{h=t}^{t+H-1} \hat{r}_h\right)^2, \qquad (6)$$

where $\hat{r}_h = \mathcal{R}\left(\hat{\phi}_{s_h}, \hat{a}_h | \theta^\mathcal{R}\right)$ is the predicted reward at time step $h$, $\hat{\phi}_{s_h} = \mathcal{M}\left(\hat{\phi}_{s_{h-1}}, \hat{a}_{h-1} | \theta^\mathcal{M}\right)$ is the latent state predicted by $\mathcal{M}$, $\hat{a}_h = \mu\left(\hat{\phi}_{s_h} | \theta^\mu\right)$ is the actor's output, and $R^*$ is the desired return. We perform $K$ gradient descent steps on $\mathcal{L}_{plan}$ (Eq. (6)) with respect to each individual action in the initial plan:

$$\hat{a}^{(i+1)}_{t:t+H-1} = \hat{a}^{(i)}_{t:t+H-1} - \alpha_{plan} \nabla_{\hat{a}^{(i)}_{t:t+H-1}} \mathcal{L}^{(i)}_{plan}, \quad 0 \leq i \leq K-1, \qquad (7)$$

where $\alpha_{plan}$ is the learning rate for plan optimization. This results in an optimal plan $\hat{a}^{(K-1)}$ whose first action is executed in the environment. Fig. 2 shows one iteration of this optimization process in which an action plan that optimizes the objective $\mathcal{L}_{plan}(\hat{a})$ given the model is inferred.

### 2.3. Intrinsically motivated meta-controller

Our approach to arbitrating between model-free and model-based control systems is based on the spatially and temporally local reliability of predictions of the latent-space world model. We define the reliability in model predictions according to the average prediction error of the model. To improve model predictions, we use the change in the average prediction error as an intrinsic reward.

#### 2.3.1. Latent-space self-organization

We incrementally self-organize the latent space into local regions with local world models using the Instantaneous Topological Map (ITM) [47] during exploration. ITM was originally designed for strongly correlated stimuli, which is the case here since the stimuli are the latent states visited along continuous trajectories, and has only a few hyperparameters. However, any other growing Self-Organizing Map (SOM) may also be used in our approach. The ITM network is defined by a set of nodes $i$, each having a weight vector $w_i$, and a set of edges connecting each node $i$ to its neighbors $N(i)$. The network starts with two connected nodes, and when a new stimulus $\phi_s$ is observed, the following adaptation steps are performed:

1. Matching: Find the nearest node $n$ and the second-nearest node $n'$ to $\phi_s$: $n \leftarrow \arg\min_i \|\phi_s - w_i\|_2^2$, $n' \leftarrow \arg\min_{j,j\neq n} \|\phi_s - w_j\|_2^2$.
2. Edge adaptation: If $n$ and $n'$ are not connected, add an edge between them. For all nodes $m \in N(n)$, if $n'$ lies inside the Thales sphere through $m$ and $n$ (the sphere with diameter $w_m w_n$), i.e. $(w_n - w_{n'})(w_m - w_{n'}) < 0$, remove the edge between $m$ and $n$, and if $m$ has no remaining edges, remove $m$.
3. Node adaptation: If $\phi_s$ lies outside the Thales sphere through $n$ and $n'$, i.e. $(w_n - \phi_s)(w_{n'} - \phi_s) > 0$, and if $\|\phi_s - w_n\|_2^2 > e_{max}$, where $e_{max}$ is the desired mapping resolution, create a new node $v$ with $w_v = \phi_s$ and an edge with $n$.

In our approach, each ITM node $n$ is assigned a local model with two neural networks $\mathcal{M}_n(.,.|\theta^{\mathcal{M}_n})$ and $\mathcal{R}_n(.,.|\theta^{\mathcal{R}_n})$ for predicting the next latent state and extrinsic reward respectively. A moving window average of model prediction error is computed and updated separately for each latent-space region $n$ (node in ITM):

$$\langle e_{t,n}^{prd} \rangle = \frac{1}{\sigma} \sum_{i=1}^{\sigma} e_i^{prd} |_{e_i^{prd} = \|\mathcal{M}_n(\phi_{s_i}, a_i|\theta^{\mathcal{M}_n}) - \phi_{s_{i+1}}\|_2^2 + \|\mathcal{R}_n(\phi_{s_i}, a_i|\theta^{\mathcal{R}_n}) - r_i^{ext}\|_2^2}, \quad (8)$$

where $\sigma$ specifies the length of the window of recent predictions in $n$, and $\mathcal{M}_n$ and $\mathcal{R}_n$ are the model's neural networks associated with $n$ for predicting the next latent state and extrinsic reward respectively. The improvement in model predictions, the change in $\langle e_{t,n}^{prd} \rangle$ over time, is then estimated by computing the learning progress (LP) locally in each region using a time window $\mathcal{W}$:

$$LP_{t,n} = \langle e_{t-\mathcal{W},n}^{prd} \rangle - \langle e_{t,n}^{prd} \rangle. \quad (9)$$

The learning progress is used to derive an intrinsic reward $r_t^{int} = -LP_{t,n}$. In order to train the model on data for which it makes large prediction errors and consequently improve the performance, this reward encourages actions that result in an increase in the average prediction error of the model. The learning progress is also used as an unbiased, spatially and temporally local reliability estimator that underlies meta-decisions, as detailed in the following section.

#### 2.3.2. Reliability-based arbitration

When a new latent state $\phi_{s_t}$ is observed, the ITM network is updated and the nearest node $n$ to $\phi_{s_t}$ is identified. If the corresponding learning progress in the latent region covered by $n$ is negative, which indicates low prediction reliability for the local world model, the meta-controller queries the model-free control system for a motor action. The model-free system in turn sends the output of the actor network $\mu(\phi_{s_t}|\theta^\mu)$ with exploration noise to the environment. If, on the other hand, the learning progress is greater or equal to zero, the meta-controller queries the model-based control system instead for a motor action. This initiates the plan optimization process (see Fig. 2). However, rather than using a predetermined planning horizon, the learning progress defined over the traversed latent regions the model-generated states belong to adaptively sets the depth of planning, as illustrated in Fig. 3. This is done by terminating the model-generated rollout when the local learning progress is negative or a maximum depth $D^{max}$ is reached (see Algorithm 1). $D^{max}$ is used to limit the computational time and has no impact on prediction reliability because the rollout is terminated before reaching $D^{max}$ if the learning progress is negative. Rolling out the model until the estimated reliability is low ensures that no imperfect model

---

**Algorithm 1** Planning-Depth ($\phi_{s_t}$, $n$, $D^{max}$)

1: $i \leftarrow t$, $\phi_{s_i} \leftarrow \phi_{s_t}$, $LP_{i,n} \leftarrow LP_{t,n}$
2: **while** ($LP_{i,n} \geq 0$) and ($i < t + D^{max}$) **do**
3:     $\hat{a}_i \leftarrow \mu(\phi_{s_i}|\theta^\mu)$, $\hat{\phi}_{s_{i+1}} \leftarrow \mathcal{M}_n(\phi_{s_i}, \hat{a}_i|\theta^{\mathcal{M}_n})$
4:     $\phi_{s_i} \leftarrow \hat{\phi}_{s_{i+1}}$
5:     $n \leftarrow$ best-matching node to $\phi_{s_i}$
6:     $i \leftarrow i + 1$
7: **end while**
8: **return** $i - t$

---

predictions are used in computing the optimal plan and reduces the computational cost. The first action of the optimal plan is then sent to the environment with exploration noise. In either case and after performing an action $a_t$, the newly collected experience $(s_t, a_t, r_t, s_{t+1})$, where $r_t = r_t^{ext} + r_t^{int}$, is added to the replay memory of recent experiences used to update the actor, critic-autoencoder, and world model networks. Fig. 4 illustrates the arbitration process of the intrinsically motivated meta-controller.

In our approach, the model-free control system provides the model-based one with a good initial action sequence. Likewise, the model-based control system provides the model-free one with a better-informed exploratory action when the model is locally reliable. Thus, the two control systems are mutually beneficial. We also limit the role of the model-based system to providing a better-informed exploratory action to the off-policy algorithm (e.g., DDPG). Our approach, therefore, improves the efficiency of learning but with less model bias than pure model-based approaches. The complete algorithm for learning visuomotor control policies with our intrinsically motivated meta-controller is given in Algorithm 2.

### 3. Integrating arbitration and imagination

In addition to planning, predictive world models can be leveraged by generating imagined experience samples to augment real-world samples and improve data efficiency of learning control policies. In a previous work, we demonstrated that performing imagined rollouts in a learned latent space and adapting the imagination depth to the improvement in learning a world model accelerate robotic visuomotor skill learning [42]. Here, we propose to integrate our learning-adaptive imagination (LA-Imagination) with the presented reliability-based arbitration using the same underlying self-organized latent space.

In LA-Imagination [42], an on-policy imagined rollout is performed every time step with a probability proportional to the local model's prediction accuracy. We modify the algorithm and instead use the adaptive-length model rollout, the input to plan optimization in our model-based control system, to provide a set of imagined transitions. To allow for learning from imagined latent-space transitions, we split the replay memory into pixel-space and latent-space replay buffers $B_{pixel}$ and $B_{latent}$ respectively. Real-world pixel-space transitions $T_i^{pixel} = (s_i, a_i, r_i, s_{i+1})$ are stored in $B_{pixel}$, while imagined latent-space transitions $T_i^{latent} = (\phi_{s_i}, \hat{a}_i, \mathcal{R}_n(\phi_{s_i}, \hat{a}_i|\theta^{\mathcal{R}_n}), \mathcal{M}_n(\phi_{s_i}, \hat{a}_i|\theta^{\mathcal{M}_n}))$ are stored in $B_{latent}$, where $\hat{a}_i = \mu(\phi_{s_i}|\theta^\mu)$, $\mathcal{M}_n$ and $\mathcal{R}_n$ are the local model networks, and $n$ is the nearest node to $\phi_{s_i}$. The learning is performed by updating parameters $\{\omega, \tilde{\omega}\}$ with gradient descent on a minibatch from $B_{pixel}$ to minimize $\mathcal{L}_{combined}$ (Eq. (4)). This is followed by updating parameters $\theta^Q$ with gradient descent using random minibatches from $B_{pixel}$ and $B_{latent}$ to minimize $\mathcal{L}_Q$ (Eq. (1)), taking the jointly optimized latent representation $\phi_{s_i}$ as input to the Q-function. Similarly, the actor parameters $\theta^\mu$ are updated by gradient descent using random minibatches from $B_{pixel}$ and $B_{latent}$ to minimize $\mathcal{L}_{DDPG}$ (Eq. (2)) or $\mathcal{L}_{CACLA}$ (Eq. (3)) according to the chosen actor–critic method (see Algorithm 3).

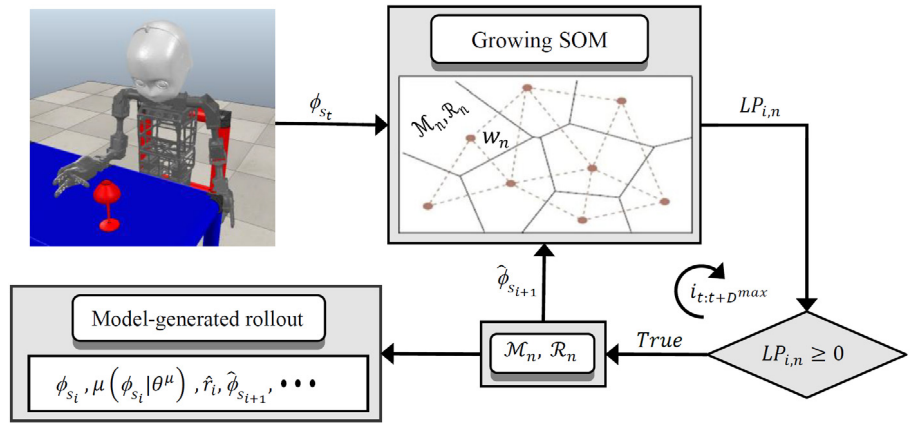

**Fig. 3.** Adaptive-length model rollout for model-based control: Given an initial latent state $\phi_{s_t}$, the world is unrolled using the model networks $\mathcal{M}_n$ and $\mathcal{R}_n$, where $n$ is the nearest node to $\phi_{s_t}$, until the learning progress in the last visited node is negative or a maximum depth $D^{max}$ is reached. At each rollout step $i$, the learning progress in the current node $n$, $LP_{i,n}$, determines whether to complete ($LP_{i,n} \geq 0$) or terminate ($LP_{i,n} < 0$) the rollout. The actions chosen during rollout are the output of the actor network $\mu(\cdot|\theta^\mu)$ and used to predict the latent states $\hat{\phi}_{s_{i+1}} = \mathcal{M}_n\left(\phi_{s_i}, \mu(\phi_{s_i}|\theta^\mu)|\theta^{\mathcal{M}_n}\right)$ and rewards $\hat{r}_i = \mathcal{R}_n\left(\phi_{s_i}, \mu(\phi_{s_i}|\theta^\mu)|\theta^{\mathcal{R}_n}\right)$. When the rollout is terminated, plan optimization is performed over the horizon $H = i - t$ (Section 2.2).

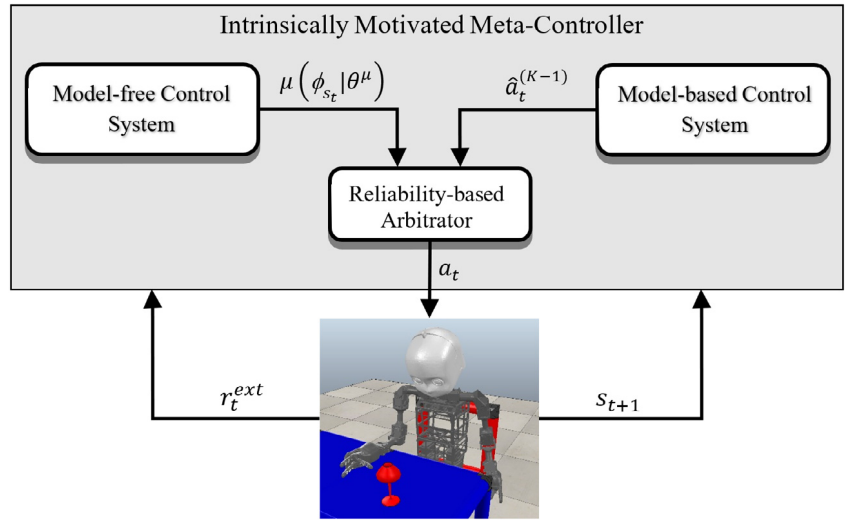

**Fig. 4.** Intrinsically Motivated Meta-Controller (IM2C): At each time step $t$, the learning progress is checked. If greater or equal to zero, the meta-controller queries the model-based control system for an action $\hat{a}_t^{(K-1)}$, which is the output of the plan optimization (Section 2.2). Otherwise, the meta-controller queries the model-free control system for an action $\mu(\phi_{s_t}|\theta^\mu)$, which is the output of the actor network. The selected action is then sent to the environment with exploration noise, and the environment returns the next state $s_{t+1}$ and extrinsic reward $r_t^{ext}$.

In our proposed framework, offline learning from imagined transitions with experience replay is coupled with online meta-control discussed in Section 2 based on the spatially and temporally local model reliability estimated by the learning progress. Fig. 5 shows the overall learning framework.

## 4. Experimental evaluation

In Sections 2 and 3, we have described the Intrinsically Motivated Meta-Controller (IM2C) and the Integrated Imagination-Arbitration (I2A) framework for improving data efficiency of learning robotic vision-based control policies. Here, we will evaluate their performance compared to baseline and state-of-the-art methods on robot grasp learning in simulation as well as on a real robot.

### 4.1. Evaluation in simulation

Here, we describe the experimental setup, including the learning parameters and robotic environment, and the results of applying our proposed and the compared algorithms to our simulated robot grasp-learning task.

#### 4.1.1. Robot grasping setup
*Parameter and implementation details.* We use the neural architectures shown in Fig. 1 with the number and size of convolutional filters placed above the corresponding layers for representing the actor and critic in the considered algorithms. No pooling layers are used. All convolutional layers are zero-padded and have stride 1. ReLU activations are used in all layers except for the output layers of the actor and critic networks that use tanh and linear activations respectively. For representing the world model, we use a fully connected neural network with one hidden layer of 20 tanh units and two output layers of 32 and 1 linear units for predicting the next latent state and extrinsic reward respectively. The weighting coefficients $\lambda_{rec}$ and $\lambda_Q$ of the combined loss function defined in Eq. (4) are set to 0.1 and 1 respectively. We set the learning rate $\alpha_{plan}$, the number of gradient descent steps $K$ and the maximum depth $D^{max}$ of the plan optimization of the model-based control system to 1e-3, 10, and 6 respectively. A

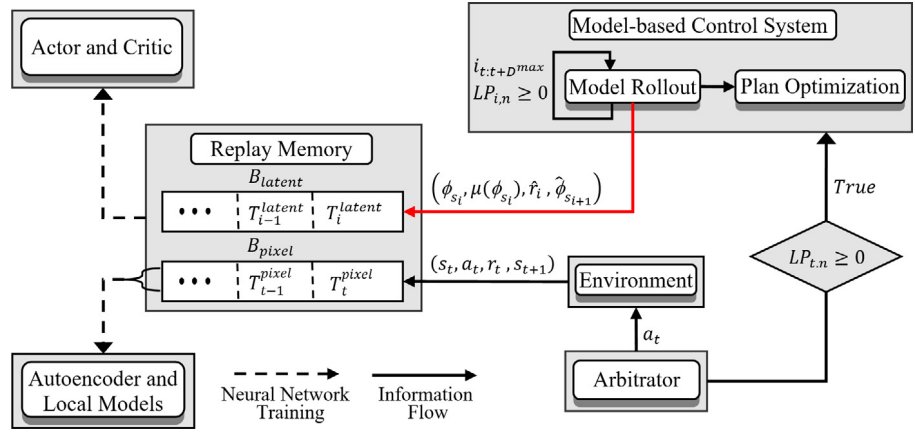

**Fig. 5.** Integrated Imagination-Arbitration (I2A) framework: At each time step $t$, the Intrinsically Motivated Meta-Controller uses the learning progress $LP_{t,n}$ associated with node $n$ to arbitrate between model-based and model-free control systems (Fig. 4). If $LP_{t,n}$ is found to be greater or equal to zero, the model-based system is called and the model is unrolled in latent space until $LP_{t,n}$ is negative or a maximum depth $D^{max}$ is reached. The resulting model rollout is used to provide a sequence of imagined transitions $(\phi_{s_i}, \mu(\phi_{s_i}), \hat{r}_i, \hat{\phi}_{s_{i+1}})$, as shown by the red arrow, which are then added to the latent-space buffer $B_{latent}$ and used to train the actor and critic networks. It is also used as input to the plan optimization process of the model-based system. After arbitration, the action of the chosen control system is sent to the environment. The collected real-world transition $(s_t, a_t, r_t, s_{t+1})$ is then stored in $B_{pixel}$ and used to train the actor, critic, and local model networks as well as the autoencoder network to jointly optimize the reconstruction and value prediction losses.

---

**Algorithm 2** Intrinsically Motivated Meta-Controller (IM2C)

---
1: **Input:** max. planning depth $D^{max}$, no. of plan optimization iterations $K$, mapping resolution $e_{max}$
2: **Given:** an off-policy actor-critic method $\mathbb{AC}$
3: Initialize learning parameters $\{\omega, \tilde{\omega}, \theta^Q, \theta^\mu, \omega', \theta^{Q'}, \theta^{\mu'}\}$
4: Initialize SOM with two nodes $n_1$ and $n_2$ and model parameters $\{\theta^{\mathcal{M}_{n_1}}, \theta^{\mathcal{R}_{n_1}}, \theta^{\mathcal{M}_{n_2}}, \theta^{\mathcal{R}_{n_2}}\}$
5: Initialize replay buffer $B$
6: **for** $episode = 1, E$ **do**
7:     Sample initial state $s_1$
8:     **for** $t = 1, T$ **do**
9:        Compute latent state encoding $\phi_{s_t} = f(s_t|\omega)$
10:       Update SOM
11:       Identify best-matching node $n$
12:       **if** $LP_{t,n} \geq 0$ **then**
13:          $H \leftarrow$ Planning-Depth $(\phi_{s_t}, n, D^{max})$      // Algorithm 1
14:          Query model-based control system with time horizon $H$    // Section 2.2
15:          $a_t \leftarrow \hat{a}_t^{(K-1)} : \hat{a}_t^{(K-1)}$ is the optimal plan's first action    // Eq. (7)
16:       **else**
17:          Query model-free control system      // Section 2.1
18:          $a_t \leftarrow \mu(\phi_{s_t}|\theta^\mu)$, where $\mu$ is $\mathbb{AC}$'s actor
19:       **end if**
20:       Add exploration noise $a_t \leftarrow a_t + \mathcal{N}(0, 1)$
21:       Execute $a_t$ and observe $r_t^{ext}$ and $s_{t+1}$
22:       Update $LP_{t,n}$, following Eq. (9), and compute intrinsic reward $r_t^{int} = -LP_{t,n}$
23:       $r_t \leftarrow r_t^{ext} + r_t^{int}$
24:       Store $(s_t, \phi_{s_t}, a_t, r_t, r_t^{ext}, s_{t+1}, \phi_{s_{t+1}})$ in $B$
25:       Update $\{\theta^{\mathcal{M}_n}, \theta^{\mathcal{R}_n}\}$ using $(\phi_{s_t}, a_t, r_t^{ext}, \phi_{s_{t+1}})$ to minimize $\mathcal{L}_{model}$     // Eq. (5)
26:       Update $\{\omega, \tilde{\omega}, \theta^Q\}$ on minibatch from $B$ to minimize $\mathcal{L}_{combined}$     // Eq. (4)
27:       Update $\theta^\mu$ on minibatch from $B$ based on the chosen $\mathbb{AC}$     // Eqs. (2) & (3)
28:       Update target network parameters: $\theta^{Q'} \leftarrow \tau \theta^Q + (1-\tau)\theta^{Q'}$,
        $\omega' \leftarrow \tau\omega + (1-\tau)\omega'$, $\theta^{\mu'} \leftarrow \tau\theta^\mu + (1-\tau)\theta^{\mu'}$ with $\tau \ll 1$
29:    **end for**
30: **end for**

---

single replay buffer with a capacity of 100k transitions is used in all experiments except for the experiment with our proposed I2A method where we use two replay buffers $B_{pixel}$ and $B_{latent}$ with capacities of 60k and 200k respectively. All networks are trained from scratch using batch size 256 and Adam optimizer [48] with learning rate 1e-3 for the critic-autoencoder and model networks and 1e-4 for the actor network. The discount factor $\gamma$ and the update rate of the target networks $\tau$ are set to 0.99 and 1e-6 respectively. The desired mapping resolution $e_{max}$ is set to 6 and the time windows used in computing the learning progress $\sigma$ and $W$ are set to 40 and 20 time units respectively. These parameters were determined empirically with respect to the best performance in terms of average episodic reward. We train the networks using Tensorflow [49] on a desktop with Intel i5-6500 CPU, 16 GB of RAM, and a single NVIDIA Geforce GTX 1050 Ti GPU.

*Simulation environment.* All experiments are conducted on our Neuro-Inspired COmpanion (NICO) robot [50] using the V-REP robot simulator [51]. NICO is a child-sized humanoid developed by the Knowledge Technology Group of the University of Hamburg. NICO is a flexible platform for research on embodied neurocognitive models based on human-like sensory and motor capabilities. It stands about one meter tall; its body proportions

**Algorithm 3** Integrated Imagination-Arbitration (I2A)

1: **Input:** max. planning depth $D^{max}$, no. of plan optimization iterations $K$, mapping resolution $e_{max}$
2: **Given:** an off-policy actor-critic method $\mathbb{AC}$
3: Initialize learning parameters $\{\omega, \tilde{\omega}, \theta^Q, \theta^\mu, \omega', \theta^{Q'}, \theta^{\mu'}\}$
4: Initialize SOM with two nodes $n_1$ and $n_2$ and model parameters $\{\theta^{\mathcal{M}_{n1}}, \theta^{\mathcal{R}_{n1}}, \theta^{\mathcal{M}_{n2}}, \theta^{\mathcal{R}_{n2}}\}$
5: Initialize replay buffers $B_{pixel}$ and $B_{latent}$
6: **for** $episode = 1, E$ **do**
7:   Sample initial state $s_1$
8:   **for** $t = 1, T$ **do**
9:     Compute latent state encoding $\phi_{s_t} = f(s_t|\omega)$
10:    Update SOM
11:    Identify best-matching node $n$
12:    **if** $LP_{t,n} \geq 0$ **then**
13:      $i \leftarrow 0$, $\phi_{s_i} \leftarrow \phi_{s_t}$, $LP_{i,n} \leftarrow LP_{t,n}$
14:      **while** $(LP_{i,n} \geq 0)$ and $(i < D^{max})$ **do**
15:        $\hat{a}_i \leftarrow \mu(\phi_{s_i}|\theta^\mu) : \mu(\cdot|\theta^\mu)$ is the actor network
16:        $\hat{r}_i \leftarrow \mathcal{R}_n(\phi_{s_i}, \hat{a}_i|\theta^{\mathcal{R}_n})$, $\hat{\phi}_{s_{i+1}} \leftarrow \mathcal{M}_n(\phi_{s_i}, \hat{a}_i|\theta^{\mathcal{M}_n})$
17:        Store imagined transition $T_i^{latent} = (\phi_{s_i}, \hat{a}_i, \hat{r}_i, \hat{\phi}_{s_{i+1}})$ in $B_{latent}$
18:        $\phi_{s_i} \leftarrow \hat{\phi}_{s_{i+1}}$, $n \leftarrow$ best-matching node to $\phi_{s_i}$
19:        $i \leftarrow i + 1$
20:      **end while**
21:      $H \leftarrow i$
22:      Query model-based control system with time horizon $H$        // Section 2.2
23:      $a_t \leftarrow \hat{a}_t^{(K-1)} : \hat{a}_t^{(K-1)}$ is the optimal plan's first action        // Eq. (7)
24:    **else**
25:      Query model-free control system        // Section 2.1
26:      $a_t \leftarrow \mu(\phi_{s_t}|\theta^\mu) : \mu(\cdot|\theta^\mu)$ is the actor network
27:    **end if**
28:    Add exploration noise $a_t \leftarrow a_t + \mathcal{N}(0, 1)$
29:    Execute $a_t$ and observe $r_t^{ext}$ and $s_{t+1}$
30:    Update $LP_{t,n}$, following Eq. (9), and compute intrinsic reward $r_t^{int} = -LP_{t,n}$
31:    $r_t \leftarrow r_t^{ext} + r_t^{int}$
32:    Store real-world transition $T_t^{pixel} = (s_t, a_t, r_t, s_{t+1})$ in $B_{pixel}$
33:    Update $\{\theta^{\mathcal{M}_n}, \theta^{\mathcal{R}_n}\}$ using $(\phi_{s_t}, a_t, r_t^{ext}, \phi_{s_{t+1}})$ to minimize $\mathcal{L}_{model}$        // Eq. (5)
34:    Fix $\theta^Q$ and update $\{\omega, \tilde{\omega}\}$ on a minibatch from $B_{pixel}$ to minimize $\mathcal{L}_{combined}$        // Eq. (4)
35:    Update $\theta^Q$ on a minibatch from $B_{pixel}$ and a minibatch from $B_{latent}$
        to minimize $\mathcal{L}_Q$, replacing $s$ with $\phi_s$        // Eq. (1)
36:    Update $\theta^\mu$ on a minibatch from $B_{pixel}$ and a minibatch from $B_{latent}$ to minimize $\mathcal{L}_{DDPG}$
        (Eq. (2)) or $\mathcal{L}_{CACLA}$ (Eq. (3)), depending on the chosen $\mathbb{AC}$ and replacing $s$ with $\phi_s$
37:    Update target network parameters: $\theta^{Q'} \leftarrow \tau\theta^Q + (1-\tau)\theta^{Q'}$,
        $\omega' \leftarrow \tau\omega + (1-\tau)\omega'$, $\theta^{\mu'} \leftarrow \tau\theta^\mu + (1-\tau)\theta^{\mu'}$ with $\tau \ll 1$
38:  **end for**
39: **end for**

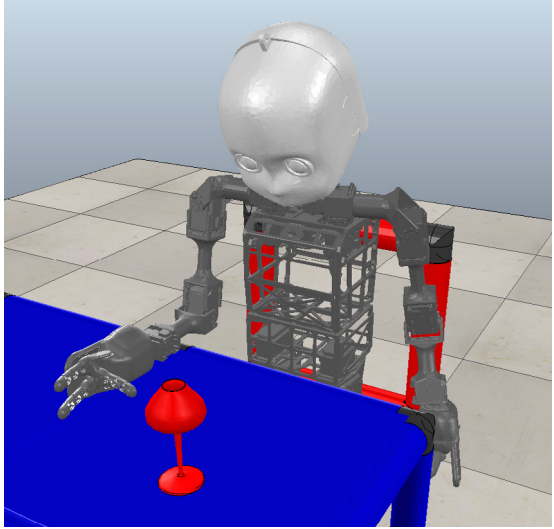

**Fig. 6.** V-REP-simulated grasp-learning scenario: NICO robot facing a table and attempting to grasp a glass randomly placed on the table.

and degrees of freedom resemble that of a three- to four-year-old child. Fig. 6 shows the configuration of the environment, including the simulated NICO robot sitting in front of a table on top of which a glass is placed and used as the grasping target.

In order to prevent self-collisions while still allowing for a large workspace, we consider learning a grasping policy that controls the shoulder joint and finger joints of the right hand, as shown in Fig. 7(a). The shoulder joint has an angular range of movement of ±100 degrees. The multi-fingered hand is tendon-operated and consists of 1 thumb and 2 index fingers with finger joints having an angular range of movement of ±160 degrees. All algorithms take as input a 64 × 32 RGB image obtained from the vision sensor, as shown in Fig. 7(b).

#### 4.1.2. Results

We run the algorithms in dense- and sparse-reward environments. Each training episode terminates when the target is grasped, toppled, or a maximum of 50 time steps is reached. The target position is randomly set to a new graspable position at the start of each episode. The extrinsic reward function is defined as follows:

$$r_t^{ext} = \begin{cases} +1 & \text{target grasped,} \\ -1 & \text{target toppled,} \\ -\|c^t - c^h\| & \text{otherwise(dense),} \\ 0 & \text{otherwise(sparse),} \end{cases}$$

where $c^t$ and $c^h$ are the center points of the target and the hand respectively. To verify successful grasps, the shoulder joint is moved 20 degrees in the opposite direction to that of the last joint position with the hand closed and the distance $\|c^t - c^h\|$ is measured. If the distance is below a threshold of 0.04 m, the grasp is deemed successful. Otherwise, the hand is opened and the shoulder joint is moved back to its last position to complete the learning episode. We compare the performance of off-policy CACLA and DDPG with and without our proposed IM2C on learning robotic vision-based grasping in dense and sparse-reward settings. Fig. 8 shows the episodic reward averaged over 5 random seeds (we use the term episodic reward to refer to the sum of extrinsic rewards collected over one complete episode). It can be observed that both CACLA+IM2C and DDPG+IM2C achieved a higher average episodic reward and a better convergence rate than their baseline counterparts at the end of training in both reward settings. The effect of IM2C is more evident in the results of learning from sparse rewards where CACLA+IM2C and DDPG+IM2C significantly outperformed their baseline counterparts in learning speed and final performance, as shown in Fig. 8(b) and Table 1. We compute the following scoring metrics: (i) Area-under-Curve (AuC) is the area under the learning curve, normalized by the total area (under Reward = 1), and gives a quantitative measure of learning speed, and (ii) Final Performance (Final Perf) is the average episodic reward over the last 500 training episodes. Results for both metrics are reported in Table 1 with no units since the two metrics are reward-based and rewards are unitless scalar values.

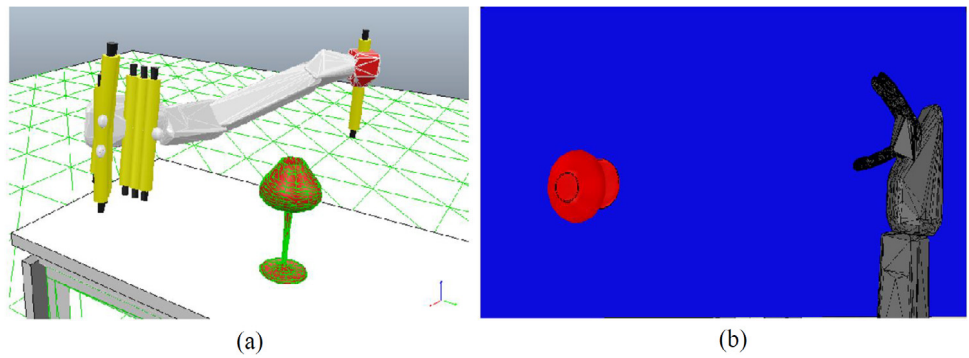

**Fig. 7.** (a) Motor output: The joints controlled by the grasping policy are depicted as yellow cylinders with one in the shoulder and 3 in each finger. (b) Sensory input: 64 × 32 RGB image used as input to the learning algorithm.

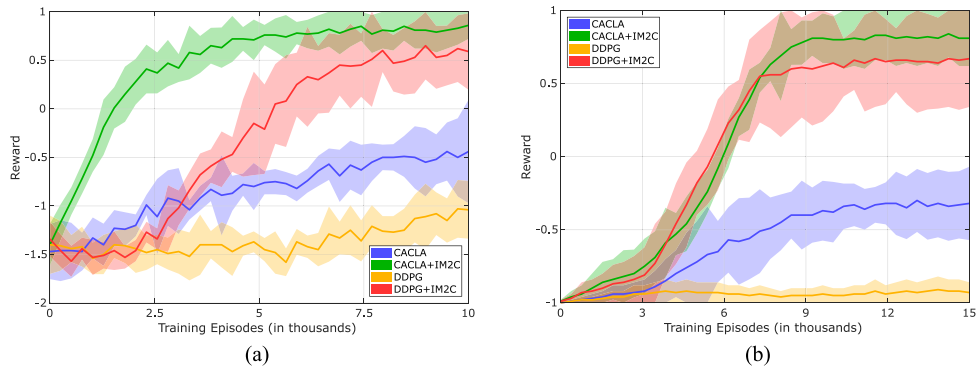

**Fig. 8.** Learning curves of off-policy CACLA and DDPG with and without IM2C on robot grasp learning from pixel input in two reward settings: (a) dense reward and (b) sparse reward. The curves are smoothed using a sliding window of 250 episodes in the dense-reward setting and 375 episodes in the sparse-reward setting. Shaded regions correspond to one standard deviation.

An empirical analysis of model learning and arbitration behavior of our proposed IM2C in the sparse-reward environment is shown in Fig. 9. We compute the mean prediction error of the model during training with DDPG+IM2C and CACLA+IM2C. Fig. 9(a) shows the model mean error, normalized to [0, 1] and averaged over 5 random seeds. As shown in the figure, the error norm of the model steadily decreased for the two algorithms. This demonstrates how the learning progress-based intrinsic reward drives the robot to constantly collect experiences that improve model predictions. The arbitration between model-free and model-based control is studied in terms of the frequency of selecting each of the two meta-control decisions per episode during training with CACLA+IM2C, averaged over 5 random seeds, as shown in Fig. 9(b). It can be observed that the behavior is predominantly model-free early in learning before the degree of control over behavior by the model-free system decreases as model predictions become more reliable. Towards the end of learning, the agent tends to use less model-free and more model-based control, with the total number of required actions (decisions) decreasing as the agent gains more experience and learns to complete the task efficiently. This emphasizes the role of the model-free system in collecting data for improving model predictions, through the proposed intrinsic reward, and the role of the model-based system in guiding behavior towards promising control policies rather than inefficient sampling of random actions by off-policy exploration.

**Table 1**
Summary statistics of the simulation results for different experimental settings.

|  | CACLA | CACLA+IM2C | DDPG | DDPG+IM2C |
|---|---|---|---|---|
| **Dense Reward** | | | | |
| AuC | 0.379 | **0.815** | 0.214 | **0.554** |
| Final Pref. | −0.5 ± 0.5 | **0.8 ± 0.1** | −1.0 ± 0.3 | **0.6 ± 0.4** |
| **Sparse Reward** | | | | |
| AuC | 0.207 | **0.579** | 0.028 | **0.548** |
| Final Pref. | −0.3 ± 0.2 | **0.8 ± 0.2** | −0.9 ± 0.1 | **0.7 ± 0.3** |

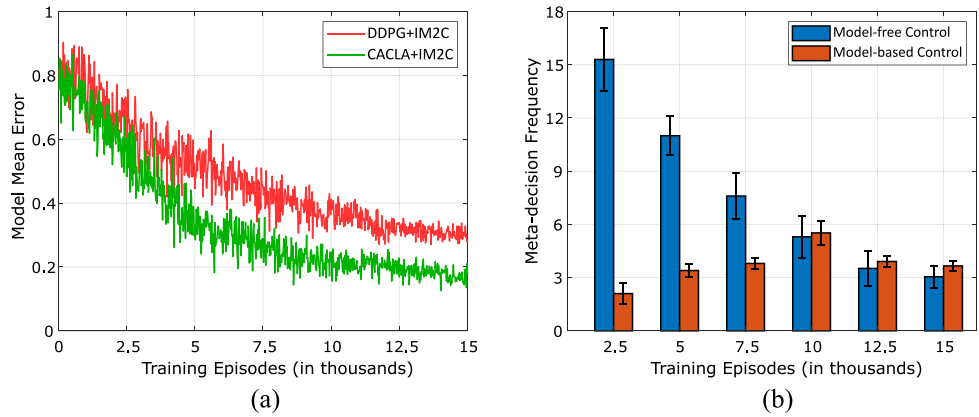

**Fig. 9.** Model learning and arbitration behavior in IM2C: (a) The mean error of model prediction per episode during training with CACLA+IM2C and DDPG+IM2C. (b) The frequency of model-free and model-based decisions (decision/episode) averaged over 2500 episodes of CACLA+IM2C.

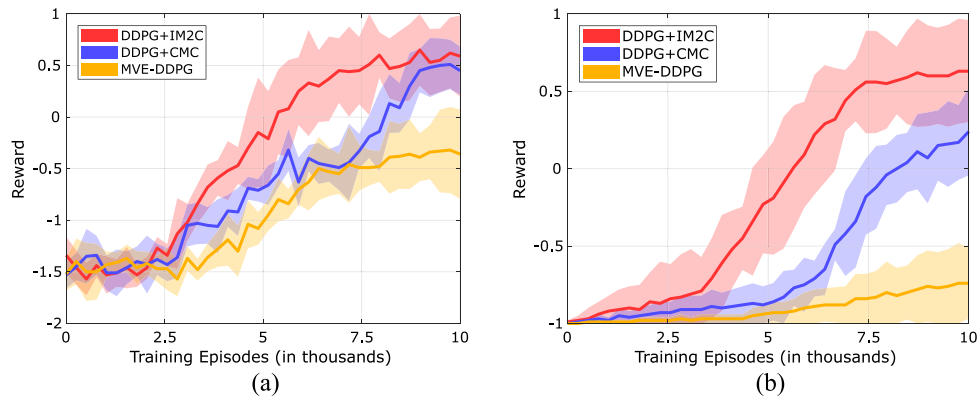

**Fig. 10.** Learning curves of DDPG+IM2C, DDPG+CMC, and MVE-DDPG on robot grasp learning from pixel input in two reward settings: (a) dense reward and (b) sparse reward. The curves are smoothed using a sliding window of 250 episodes. Shaded regions correspond to one standard deviation.

We also compare IM2C to previous methods for improving model-free value estimation with model-based predictions, particularly the state-of-the-art Model-based Value Expansion (MVE) method [28] and the more recent Curious Meta-Controller (CMC) method [33]. We implement MVE-DDPG from [28] and DDPG+CMC from [33] with a prediction horizon $H$ of 2 and 3 steps respectively, and find these values to produce the best results. Fig. 10 shows the average episodic reward over 5 random seeds. The three methods have a comparable learning performance over 3k episodes in the dense-reward setting (Fig. 10(a)). The episodic reward of DDPG+IM2C and DDPG+CMC, however, continues to increase faster than that of MVE-DDPG, reaching 0.59 and 0.45 respectively. In the sparse-reward setting (Fig. 10(b)), MVE-DDPG shows no clear improvement in performance, while DDPG+IM2C and DDPG+CMC are able to improve their performance, converging to a policy of 0.62 and 0.24 episodic reward respectively. We believe the poor performance of MVE is primarily due to incorporating imperfect predictions in learning value estimates, as opposed to the reliability-driven model use of IM2C. Besides, CMC and MVE use fixed $H$, increasing the risk of compounding prediction errors, while IM2C enables automatic selection of $H$ that is fully adaptive to the local reliability of the model.

Last but not least, we evaluate our proposed I2A framework, which combines experience imagination with reliability-based arbitration, by conducting an ablation study to analyze the influence of individual components of I2A, namely the arbitration and imagination components. This is performed by comparing I2A to IM2A that represents the arbitration component and to LA-Imagination (see Section 3) that represents the imagination component. The average episodic reward of running the three algorithms over 5 random seeds in the two reward settings is shown in Fig. 11. It is clear that augmenting the replay memory of DDPG with latent-space imagined transitions using LA-Imagination significantly improves the data efficiency of DDPG which otherwise completely fails to show any progress (see Fig. 8). Compared to DDPG+LA-Imagination, DDPG+IM2C leads to a higher episodic reward, which again confirms the effectiveness of the meta-controller in adaptively arbitrating between model-based and model-free control systems and choosing more informed exploratory actions, progressing faster to a good grasping policy. DDPG+I2A, on the other hand, yields the best results through combining the advantages of the two approaches using the same underlying self-organized latent space.

### 4.2. Evaluation on a real robot

For the experiments with the physical NICO, the simulation environment was recreated as faithfully as possible: The simulation is based on a URDF model of NICO. Therefore, there is no difference in the simulated and the real robot. Both the table and NICO's seat have the same height as in the simulation, allowing for a direct transfer of the arm pose and, more importantly, the trained neural model. Furthermore, the color of the table is identical to the color in the simulation. A grasping object is slightly different in geometry, to allow for more stable grasps, but has the same color. To achieve the same perspective for the visual input, an external camera was mounted on the table with a view similar to the simulated camera. Fig. 12 shows NICO in the experimental setup.

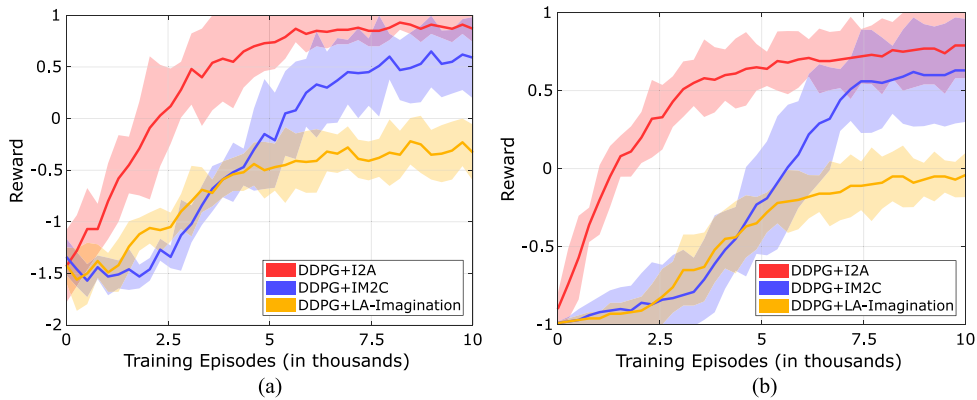

**Fig. 11.** Learning curves of DDPG+I2A, DDPG+IM2C, and DDPG+LA-Imagination on robot grasp learning from pixel input in two reward settings: (a) dense reward and (b) sparse reward. The curves are smoothed using a sliding window of 250 episodes. Shaded regions correspond to one standard deviation.

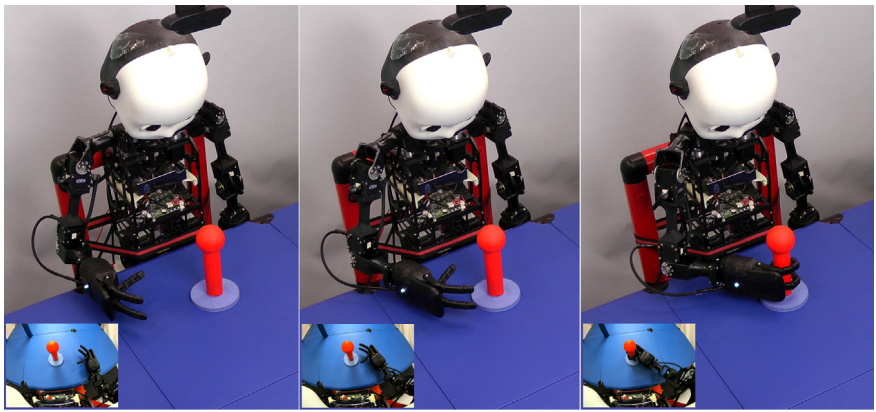

**Fig. 12.** NICO experimental setup during a grasping test trial. From left to right: the exocentric and the egocentric (inset) views of the initial, intermediate, and full-grasp configurations.

While the grasping object's position in the simulation environment can be manipulated directly and the virtual NICO is only used for grasp learning and execution, NICO in the real environment is also used to place the object at an exact and known position on the table. Each grasping trial consists of the following steps: Starting from the initial position (shoulder at zero degrees), the grasping object is put into NICO's hand (if it was not already in the hand), the hand closes and NICO puts the object at a predetermined position on the table, the position is memorized and NICO moves the hand back to the starting position. Now the actual grasping trial starts by taking an image with the external camera and feeding it into the actor network that outputs a motor command to NICO. After the movement is executed, either the object is grasped, or if the hand is too far from the object, another image is recorded, and the process is repeated. Up to eight consecutive grasping steps are performed before the attempt is categorized as failed, and the object is retrieved using its initially stored position.

To compare the performance of the algorithms on the real NICO robot, we take the best-performing policy network of each algorithm, trained in simulation, and then deploy it on the real robot. We perform 25 test episodes, each with a random graspable position. To achieve a seamless simulation-to-real transfer of the trained policy networks and to compensate for the slightly different alignment of the simulated and the real camera, we force the encoder part of the critic-autoencoder network to map one image from the simulation environment and one image from the real world with the same joint configuration and environmental setup into a similar latent representation. This is done by training the encoder to minimize the loss $\mathcal{L}_{Transfer} = \frac{1}{2} \|\phi_{s_{sim}} - \phi_{s_{real}}\|_2^2$ over a training set of 2k simulated-real image pairs (for 200 epochs with batch size 512), as described in Fig. 13. The encoder then computes the latent state to be used as input to the policy network during real-world testing. No fine-tuning of the trained policy networks is performed. We report the success rate (the proportion of the successful test episodes) for each algorithm in Table 2.

**Table 2**
Success rate of the trained policy networks on the real robot.

| Environment | DDPG | DDPG+CMC | MVE-DDPG | DDPG+IM2C | DDPG+I2A |
|---|---|---|---|---|---|
| Dense reward | 16% | 68% | 48% | 80% | **88%** |
| Sparse reward | 12% | 44% | 12% | **76%** | **76%** |

## 5. Conclusion

We presented a novel robot dual-system motor learning approach that is behaviorally and neurally plausible, data efficient, and competitive with the state of the art. Our approach adaptively arbitrates between model-based and model-free decisions based on the spatially and temporally local reliability of a learned world model. The reliability estimate computed locally for every region of a learned latent space is used to make the meta-decision as well as to enable an adaptive-length model rollout for plan optimization during model-based control. We derive an intrinsic reward using the reliability estimate to encourage collecting experience data that improves the model. To further improve the data efficiency, we leverage the reliable multi-step model predictions by combining arbitration with experience imagination where imagined experiences collected from model rollouts are used as an additional training data for the control policy.

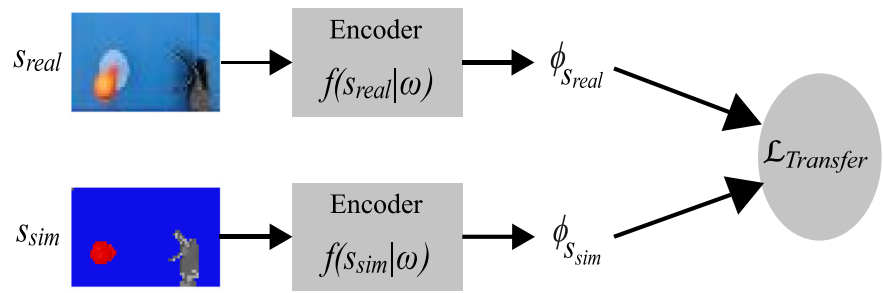

**Fig. 13.** Simulation-to-real transfer procedure. The encoder $f(\cdot|\omega)$, trained in simulation during grasp learning, is refined with supervised learning to map 64 × 32 RGB simulated $s_{sim}$ and real-world $s_{real}$ images related to the same environmental and task settings into a similar latent representation. Both real and simulated images are the output of a top-mounted camera.

We show that our approach learns better vision-based control policies than baseline and state-of-the-art methods in dense and sparse reward environments. Policy networks trained in simulation with our approach are shown to perform well on the physical robot without fine-tuning of the policy parameters. Our results suggest that model reliability is essential for dual-system approaches involving online meta-decisions to determine which of the model-based and model-free systems to query for an action and for generating imagined experience data that includes less overall prediction error. Our approach can be used with any off-policy reinforcement learning algorithm, which we demonstrated with off-policy CACLA and DDPG. We believe that our approach can be extended to the case of a multi-step model, instead of the single-step model used in the present work, by incorporating temporal abstractions, such as options [52,53]. Another promising direction for future work is to generalize our approach to environments with stochastic dynamics.

**Declaration of competing interest**

The authors declare that they have no known competing financial interests or personal relationships that could have appeared to influence the work reported in this paper.

**Acknowledgments**

This work was supported by the German Academic Exchange Service (DAAD) funding programme (No. 57214224) with partial support from the German Research Foundation DFG under project CML (TRR 169).

**References**

[1] V. Mnih, K. Kavukcuoglu, D. Silver, A.A. Rusu, J. Veness, M.G. Bellemare, A. Graves, M. Riedmiller, A.K. Fidjeland, G. Ostrovski, et al., Human-level control through deep reinforcement learning, Nature 518 (7540) (2015) 529–533.
[2] D. Silver, A. Huang, C.J. Maddison, A. Guez, L. Sifre, G. Van Den Driessche, J. Schrittwieser, I. Antonoglou, V. Panneershelvam, M. Lanctot, et al., Mastering the game of Go with deep neural networks and tree search, Nature 529 (7587) (2016) 484.
[3] S. Levine, C. Finn, T. Darrell, P. Abbeel, End-to-end training of deep visuomotor policies, J. Mach. Learn. Res. 17 (1) (2016) 1334–1373.
[4] S. Gu, E. Holly, T. Lillicrap, S. Levine, Deep reinforcement learning for robotic manipulation with asynchronous off-policy updates, in: 2017 IEEE International Conference on Robotics and Automation, ICRA, 2017, pp. 3389–3396.
[5] T. Schaul, J. Quan, I. Antonoglou, D. Silver, Prioritized experience replay, in: 4th International Conference on Learning Representations, 2016.
[6] M. Andrychowicz, F. Wolski, A. Ray, J. Schneider, R. Fong, P. Welinder, B. McGrew, J. Tobin, O.P. Abbeel, W. Zaremba, Hindsight experience replay, in: Advances in Neural Information Processing Systems, 2017, pp. 5048–5058.
[7] J. Fu, J. Co-Reyes, S. Levine, Ex2: Exploration with exemplar models for deep reinforcement learning, in: Advances in Neural Information Processing Systems, 2017, pp. 2577–2587.
[8] H. Tang, R. Houthooft, D. Foote, A. Stooke, O.X. Chen, Y. Duan, J. Schulman, F. DeTurck, P. Abbeel, # exploration: A study of count-based exploration for deep reinforcement learning, in: Advances in Neural Information Processing Systems, 2017, pp. 2753–2762.
[9] T. Xu, Q. Liu, L. Zhao, J. Peng, Learning to explore via meta-policy gradient, in: International Conference on Machine Learning, 2018, pp. 5463–5472.
[10] F. Garcia, P.S. Thomas, A meta-mdp approach to exploration for lifelong reinforcement learning, in: Advances in Neural Information Processing Systems, 2019, pp. 5692–5701.
[11] D. Pathak, P. Agrawal, A.A. Efros, T. Darrell, Curiosity-driven exploration by self-supervised prediction, in: Proceedings of the 34th International Conference on Machine Learning, Vol. 70, 2017, pp. 2778–2787.
[12] Y. Burda, H. Edwards, A. Storkey, O. Klimov, Exploration by random network distillation, in: 7th International Conference on Learning Representations, 2019.
[13] J. Gottlieb, P.-Y. Oudeyer, M. Lopes, A. Baranes, Information-seeking, curiosity, and attention: computational and neural mechanisms, Trends Cogn. Sci. 17 (11) (2013) 585–593.
[14] A. Péré, S. Forestier, O. Sigaud, P.-Y. Oudeyer, Unsupervised learning of goal spaces for intrinsically motivated goal exploration, in: 6th International Conference on Learning Representations, 2018.
[15] F. Mannella, V.G. Santucci, E. Somogyi, L. Jacquey, K.J. O'Regan, G. Baldassarre, Know your body through intrinsic goals, Front. Neurorobot. 12 (2018) 30.
[16] S. Mohamed, D.J. Rezende, Variational information maximisation for intrinsically motivated reinforcement learning, in: Advances in Neural Information Processing Systems, 2015, pp. 2125–2133.
[17] E. Hazan, S. Kakade, K. Singh, A. Van Soest, Provably efficient maximum entropy exploration, in: International Conference on Machine Learning, 2019, pp. 2681–2691.
[18] A.M. Haith, J.W. Krakauer, Model-based and model-free mechanisms of human motor learning, in: Progress in Motor Control, Springer, 2013, pp. 1–21.
[19] S.W. Lee, S. Shimojo, J.P. O'Doherty, Neural computations underlying arbitration between model-based and model-free learning, Neuron 81 (3) (2014) 687–699.
[20] N.D. Daw, Y. Niv, P. Dayan, Uncertainty-based competition between prefrontal and dorsolateral striatal systems for behavioral control, Nature Neurosci. 8 (12) (2005) 1704–1711.
[21] F. Cushman, A. Morris, Habitual control of goal selection in humans, Proc. Natl. Acad. Sci. 112 (45) (2015) 13817–13822.
[22] M. Keramati, P. Smittenaar, R.J. Dolan, P. Dayan, Adaptive integration of habits into depth-limited planning defines a habitual-goal–directed spectrum, Proc. Natl. Acad. Sci. 113 (45) (2016) 12868–12873.
[23] W. Kool, S.J. Gershman, F.A. Cushman, Planning complexity registers as a cost in metacontrol, J. Cogn. Neurosci. 30 (10) (2018) 1391–1404.
[24] Y.-L. Boureau, P. Sokol-Hessner, N.D. Daw, Deciding how to decide: Self-control and meta-decision making, Trends Cogn. Sci. 19 (11) (2015) 700–710.
[25] F. Lieder, T.L. Griffiths, When to use which heuristic: A rational solution to the strategy selection problem, in: Proceedings of the 37th Annual Conference of the Cognitive Science Society, 2015.

[26] E.M. Russek, I. Momennejad, M.M. Botvinick, S.J. Gershman, N.D. Daw, Predictive representations can link model-based reinforcement learning to model-free mechanisms, PLoS Comput. Biol. 13 (9) (2017) e1005768.
[27] A. Nagabandi, G. Kahn, R.S. Fearing, S. Levine, Neural network dynamics for model-based deep reinforcement learning with model-free fine-tuning, in: 2018 IEEE International Conference on Robotics and Automation, ICRA, 2018, pp. 7559–7566.
[28] V. Feinberg, A. Wan, I. Stoica, M.I. Jordan, J.E. Gonzalez, S. Levine, Model-based value estimation for efficient model-free reinforcement learning, 2018, arXiv preprint arXiv:1803.00101.
[29] S. Racanière, T. Weber, D. Reichert, L. Buesing, A. Guez, D.J. Rezende, A.P. Badia, O. Vinyals, N. Heess, Y. Li, et al., Imagination-augmented agents for deep reinforcement learning, in: Advances in Neural Information Processing Systems, 2017, pp. 5690–5701.
[30] D. Ha, J. Schmidhuber, World models, 2018, arXiv preprint arXiv:1803.10122.
[31] V. François-Lavet, Y. Bengio, D. Precup, J. Pineau, Combined reinforcement learning via abstract representations, in: Proceedings of the AAAI Conference on Artificial Intelligence, Vol. 33, 2019, pp. 3582–3589.
[32] F.S. Fard, T.P. Trappenberg, Mixing habits and planning for multi-step target reaching using arbitrated predictive actor-critic, in: 2018 International Joint Conference on Neural Networks, IJCNN, 2018, pp. 1–8.
[33] M.B. Hafez, C. Weber, M. Kerzel, S. Wermter, Curious meta-controller: Adaptive alternation between model-based and model-free control in deep reinforcement learning, in: 2019 International Joint Conference on Neural Networks, IJCNN, 2019, pp. 1–8.
[34] S.T. Moulton, S.M. Kosslyn, Imagining predictions: mental imagery as mental emulation, Philos. Trans. R. Soc. B 364 (1521) (2009) 1273–1280.
[35] L.K. Case, J. Pineda, V.S. Ramachandran, Common coding and dynamic interactions between observed, imagined, and experienced motor and somatosensory activity, Neuropsychologia 79 (2015) 233–245.
[36] R. Ptak, A. Schnider, J. Fellrath, The dorsal frontoparietal network: a core system for emulated action, Trends Cogn. Sci. 21 (8) (2017) 589–599.
[37] J.E. Driskell, C. Copper, A. Moran, Does mental practice enhance performance? J. Appl. Psychol. 79 (4) (1994) 481.
[38] S. Mahadevan, Imagination machines: a new challenge for artificial intelligence, in: Thirty-Second AAAI Conference on Artificial Intelligence, 2018.
[39] J.B. Hamrick, Analogues of mental simulation and imagination in deep learning, Curr. Opin. Behav. Sci. 29 (2019) 8–16.
[40] S. Gu, T. Lillicrap, I. Sutskever, S. Levine, Continuous deep q-learning with model-based acceleration, in: International Conference on Machine Learning, pp. 2829–2838.
[41] G. Kalweit, J. Boedecker, Uncertainty-driven imagination for continuous deep reinforcement learning, in: Conference on Robot Learning, 2017, pp. 195–206.
[42] M.B. Hafez, C. Weber, M. Kerzel, S. Wermter, Efficient intrinsically motivated robotic grasping with learning-adaptive imagination in latent space, in: 2019 Joint IEEE 9th International Conference on Development and Learning and Epigenetic Robotics, ICDL-EpiRob, 2019, pp. 240–246.
[43] B. Amos, I. Jimenez, J. Sacks, B. Boots, J.Z. Kolter, Differentiable MPC for end-to-end planning and control, in: Advances in Neural Information Processing Systems, 2018, pp. 8289–8300.
[44] N.R. Ke, A. Singh, A. Touati, A. Goyal, Y. Bengio, D. Parikh, D. Batra, Learning dynamics model in reinforcement learning by incorporating the long term future, in: 7th International Conference on Learning Representations, 2019.
[45] T.P. Lillicrap, J.J. Hunt, A. Pritzel, N. Heess, T. Erez, Y. Tassa, D. Silver, D. Wierstra, Continuous control with deep reinforcement learning, in: 4th International Conference on Learning Representations, 2016.
[46] H. Van Hasselt, Reinforcement learning in continuous state and action spaces, in: Reinforcement Learning, Springer, 2012, pp. 207–251.
[47] J. Jockusch, H. Ritter, An instantaneous topological mapping model for correlated stimuli, in: International Joint Conference on Neural Networks, Vol. 1, IJCNN, 1999, pp. 529–534.
[48] D.P. Kingma, J. Ba, Adam: A method for stochastic optimization, in: 3rd International Conference on Learning Representations, 2015.
[49] M. Abadi, P. Barham, J. Chen, Z. Chen, A. Davis, J. Dean, M. Devin, S. Ghemawat, G. Irving, M. Isard, et al., Tensorflow: A system for large-scale machine learning, in: 12th {USENIX} Symposium on Operating Systems Design and Implementation, {OSDI} 16, 2016, pp. 265–283.
[50] M. Kerzel, E. Strahl, S. Magg, N. Navarro-Guerrero, S. Heinrich, S. Wermter, NICO–neuro-inspired companion: A developmental humanoid robot platform for multimodal interaction, in: 2017 26th IEEE International Symposium on Robot and Human Interactive Communication, RO-MAN, 2017, pp. 113–120.
[51] E. Rohmer, S.P. Singh, M. Freese, V-REP: A versatile and scalable robot simulation framework, in: 2013 IEEE/RSJ International Conference on Intelligent Robots and Systems, 2013, pp. 1321–1326.
[52] R.S. Sutton, D. Precup, S. Singh, Between MDPs and semi-MDPs: A framework for temporal abstraction in reinforcement learning, Artificial Intelligence 112 (1–2) (1999) 181–211.
[53] D. Precup, Temporal Abstraction in Reinforcement Learning (Ph.D. thesis), University of Massachusetts, 2000.

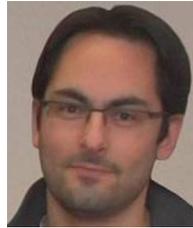

**Muhammad Burhan Hafez** received the M.Sc. degree in computer science from the University of Malaya, Kuala Lumpur, Malaysia, and the Ph.D. degree in computer science from Universität Hamburg, Hamburg, Germany. He is currently a postdoctoral associate at the Knowledge Technology Group, Universität Hamburg, Hamburg, Germany. His research interests include deep reinforcement learning, intrinsic motivation, meta-decision making, and cognitive robotics.

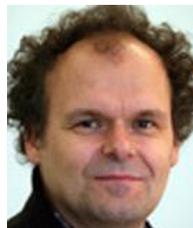

**Cornelius Weber** received the Diploma degree in physics, from the University of Bielefeld, Bielefeld, Germany, and the Ph.D. degree in computer science with the Technische Universität Berlin, Berlin, Germany, in 2000. He is a Laboratory Manager with the Knowledge Technology Group, Universität Hamburg, Hamburg, Germany. He was a Post-Doctoral Fellow of Brain and Cognitive Sciences with the University of Rochester, Rochester, NY, USA. From 2002 to 2005, he was a Research Scientist of Hybrid Intelligent Systems with the University of Sunderland, Sunderland, U.K. He was a Junior Fellow with the Frankfurt Institute for Advanced Studies, Frankfurt am Main, Germany, until 2010. His current research interests include computational neuroscience with a focus on vision, unsupervised learning, and reinforcement learning.

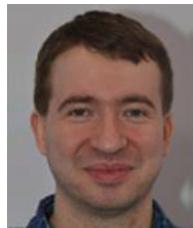

**Matthias Kerzel** received his Diploma (German M.Sc.) and Ph.D. in computer science from the Universität Hamburg, Germany. He currently works as a post-doctoral research associate at Knowledge Technology, Universität Hamburg in the context of the international collaborative research centre Crossmodal Learning (TRR-169). His research interests include artificial intelligence and developmental, humanoid robotics. In these areas, he currently focuses on neurocognitive models for deep reinforcement learning and the autonomous development of interactive multimodal perception and sensorimotor abilities through interaction with the environment.

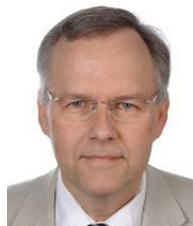

**Stefan Wermter** is Full Professor at the University of Hamburg, Germany, and Director of the Knowledge Technology Institute in the Dept. of Informatics. His main research interests are in the fields of neural networks, hybrid knowledge technology, cognitive robotics, and human-robot interaction. He has been an associate editor of the journals 'Transactions on Neural Networks and Learning Systems', and is an associate editor of 'Connection Science' and 'International Journal for Hybrid Intelligent Systems' and he is on the editorial board of the journals 'Cognitive Systems Research', 'Cognitive Computation' and 'Journal of Computational Intelligence'. Currently, he is co-coordinator of the international collaborative research centre on Crossmodal Learning (TRR-169) and coordinator of the European Training Network SECURE on safety for cognitive robots. In 2019 he has been elected as the President for the European Neural Network Society 2020-2022.